\documentclass[10pt,a4paper,conference]{IEEEtran}
\usepackage{cite}
\usepackage{amsmath,amssymb,amsfonts}
\usepackage{algorithmic}
\usepackage{graphicx}
\usepackage{textcomp}
\usepackage{xcolor}

\usepackage{subfigure}
\usepackage{dsfont}
\usepackage{multirow}
\usepackage[ruled, norelsize]{algorithm2e}

\usepackage[marginal]{footmisc}

\ifCLASSINFOpdf
\else
\fi
\begin{document}
%
\title{Adaptive Matching of Kernel Means}

\author{\IEEEauthorblockN{Miao Cheng\textsuperscript{$ \dagger $}
}
\IEEEauthorblockA{School of Computer Science and Information Engineering\\
Guangxi Normal University\\
Guilin, Guangxi, China\\
Email: miao\_cheng@outlook.com}
\and
\IEEEauthorblockN{Xinge You}
\IEEEauthorblockA{School of Electronic Information and Communications\\
Huazhong University of Science and Technology\\
Wuhan, Hubei, China\\
Email: youxg@hust.edu.cn}
}


%


\maketitle

\begin{abstract}
\footnote{\textsuperscript{$ \dagger $} Corresponding author. This work was partly supported by Innovation and Talent Foundation of Guangxi Province (RZ1900007485).}As a promising step, the performance of data analysis and feature learning are able to be improved if certain pattern matching mechanism is available. One of the feasible solutions can refer to the importance estimation of instances, and consequently, kernel mean matching (KMM) has become an important method for knowledge discovery and novelty detection in kernel machines. 
Furthermore, the existing KMM methods have focused on concrete learning frameworks.
In this work, a novel approach to adaptive matching of kernel means is proposed, and selected data with high importance are adopted to achieve calculation efficiency with optimization. In addition, scalable learning can be conducted in proposed method as a generalized solution to matching of appended data. The experimental results on a wide variety of real-world data sets demonstrate the proposed method is able to give outstanding performance compared with several state-of-the-art methods, while calculation efficiency can be preserved.
\end{abstract}


%
\IEEEpeerreviewmaketitle

\section{Introduction}
Knowledge understanding and management are the key technologies in information systems, and many progresses have been developed to meet increasing demands of data analysis and handling.  
In traditional pattern analysis methods, it is not guaranteed the ideal fact that available information of training samples is suitable with testing ones completely \cite{Bickel09CovShift}\cite{Huang07CSSB}\cite{Zadrozny04SSB}. An existing challenge refers to knowledge transfer between data sets belong to different groups
\cite{Gretton12TST}\cite{Cheng15CRH}. 
Similarly, knowledge matching is to exploit the similarities of different sets, and estimate the distribution of data to improve performance of information systems \cite{Candela08DSML}.

In the context of importance sampling, the \emph{importance} indicates the ratio of two probability density functions \cite{Shimodaira00CSWLLF}. And there are two main categories of tasks such as covariate shift adaptation \cite{Sugiyama05IECS}\cite{Sugiyama07EGECS} or outlier detection \cite{Scholkopf01ES}\cite{Tax04SVDD}.
It is believed that, the densities of data are quite related with the importance of instances \cite{Bishop11PRML}\cite{Hastie11ESL}.
Until now, there have been lots of available approaches to calculate densities of data, and most of them can be absorbed into a generalized learning framework of gaussian kernels \cite{Sugiyama07DIE}\cite{Yu12KMM}. 
Furthermore, explicit calculations of densities can be avoided.
Besides, kernel learning is also considered in the nonlinear embedding with optimal distribution of positive definite kernels \cite{Tolstikhin17KME}\cite{Cheng18MMC}.
Suppose that matching input samples $ { x }_{ i }^{ m }, \left( i=1,2,\cdots ,{ n }_{ m } \right) $ from a matching input distribution with density $ p_{m} \left( x \right) $ and i.i.d. reference input samples $ { x }_{ i }^{ r }, \left( i=1,2,\cdots ,{ n }_{ r } \right) $ from a reference input distribution with density $ p_{r} \left( x \right) $. 
Without loss of generality, the importance of a given sample $ w \left( x \right) $ \cite{Sugiyama07DIE} is given by the ratio of densities $ p_{r}  \left( x \right) $ and $ p_{m} \left( x \right) $ as
\begin{equation}
  w\left( x \right) =\frac { { p }_{ r }\left( x \right)  }{ { p }_{ m }\left( x \right)  } .
\end{equation}

In the literature, the most popular solutions have referred to kernel mean matching (KMM) \cite{Yu12KMM}\cite{Miao15enKMM}\cite{Miao15LADR}, which employs kernel functions to approximately describe the distributions of data blocks \cite{Scholkopf01Kernels}\cite{Cristianini01KTA}\cite{Cortes12CA}\cite{Cortes17SKM}.
Usually, it is reduced to be a kernel learning problem, and importance of samples can be estimated for further selection or learning. For example, 
the sample importance can be modeled using Kullback-Leibler divergence \cite{Sugiyama07DIE}, and a least-square approach is proposed to direct  importance estimation \cite{Kanamori09LSDIE}, which is the most popular framework till now. Particularly, an ensemble learning method is designed to handle KMM problem by dividing reference samples into smaller partitions, and density ratio of each partition is calculated to  be fused with a weighted sum \cite{Miao15enKMM}\cite{Chandra16SKMM}. In addition, a locally adaptive kernel is introduced to estimation of density ratio for kernel mean matching \cite{Miao15LADR}. 
In this work, a novel approach to adaptive matching of kernel means (AMKM) is devised for efficient matching. 
By selecting the top important instances, the proposed method can significantly improve calculation complexity. It is shown that, the proposed method are able to get outstanding results compared with the state-of-the-art KMM methods, while calculation efficiency can be achieved. Furthermore, scalable KMM can be conducted in the proposed method for incremental reference instances of mean matching.

In the rest of this paper, it is organized as follows. Firstly, some background of KMM is introduced in Section II. The details of proposed method is given in Section III. The experimental results are given in Section IV. Finally, the conclusion is draw in Section V.

\begin{figure*}
    \centering
    \subfigure[]{ \includegraphics[width=0.31\textwidth]{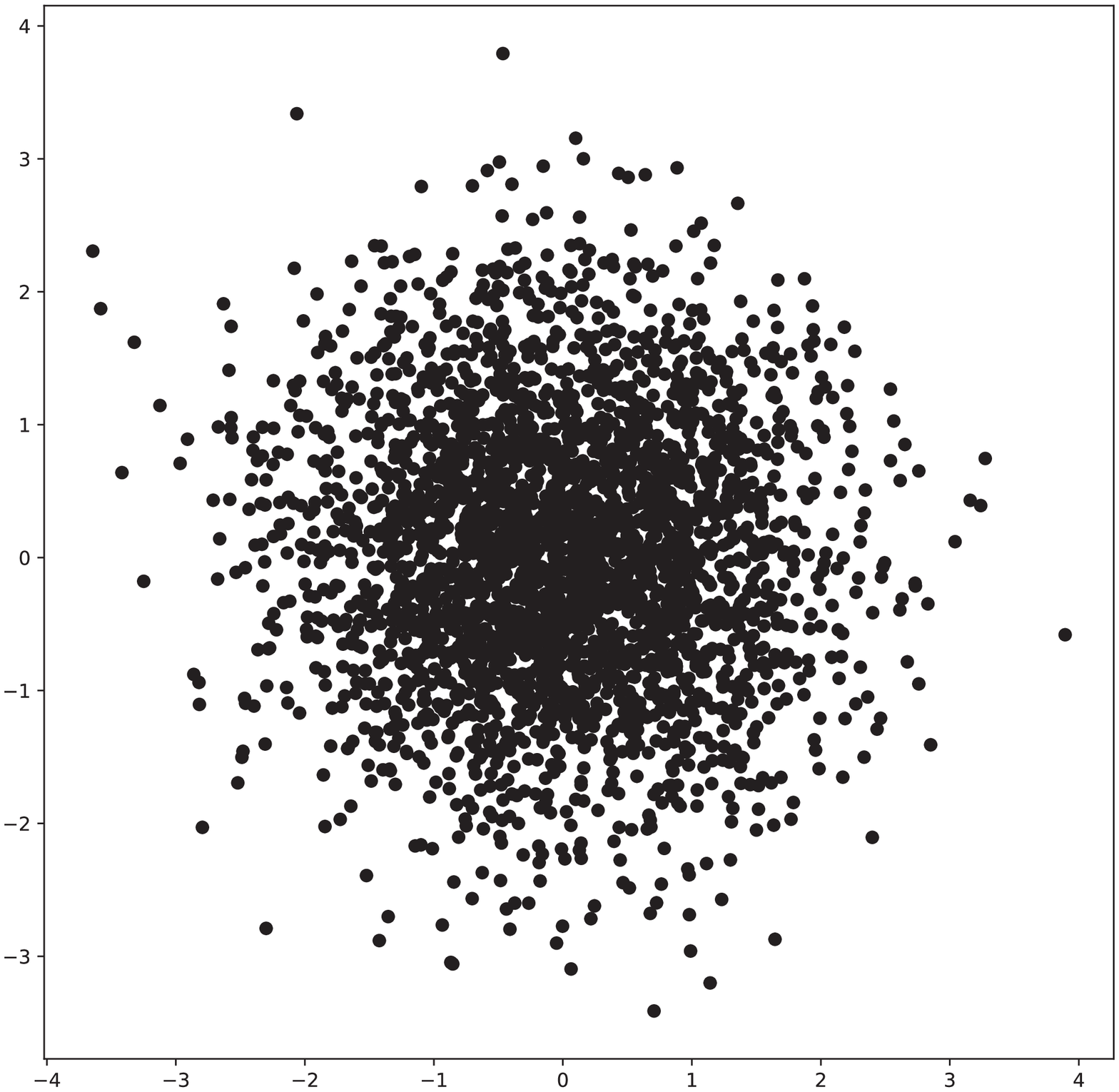}	}
    \subfigure[]{ \includegraphics[width=0.31\textwidth]{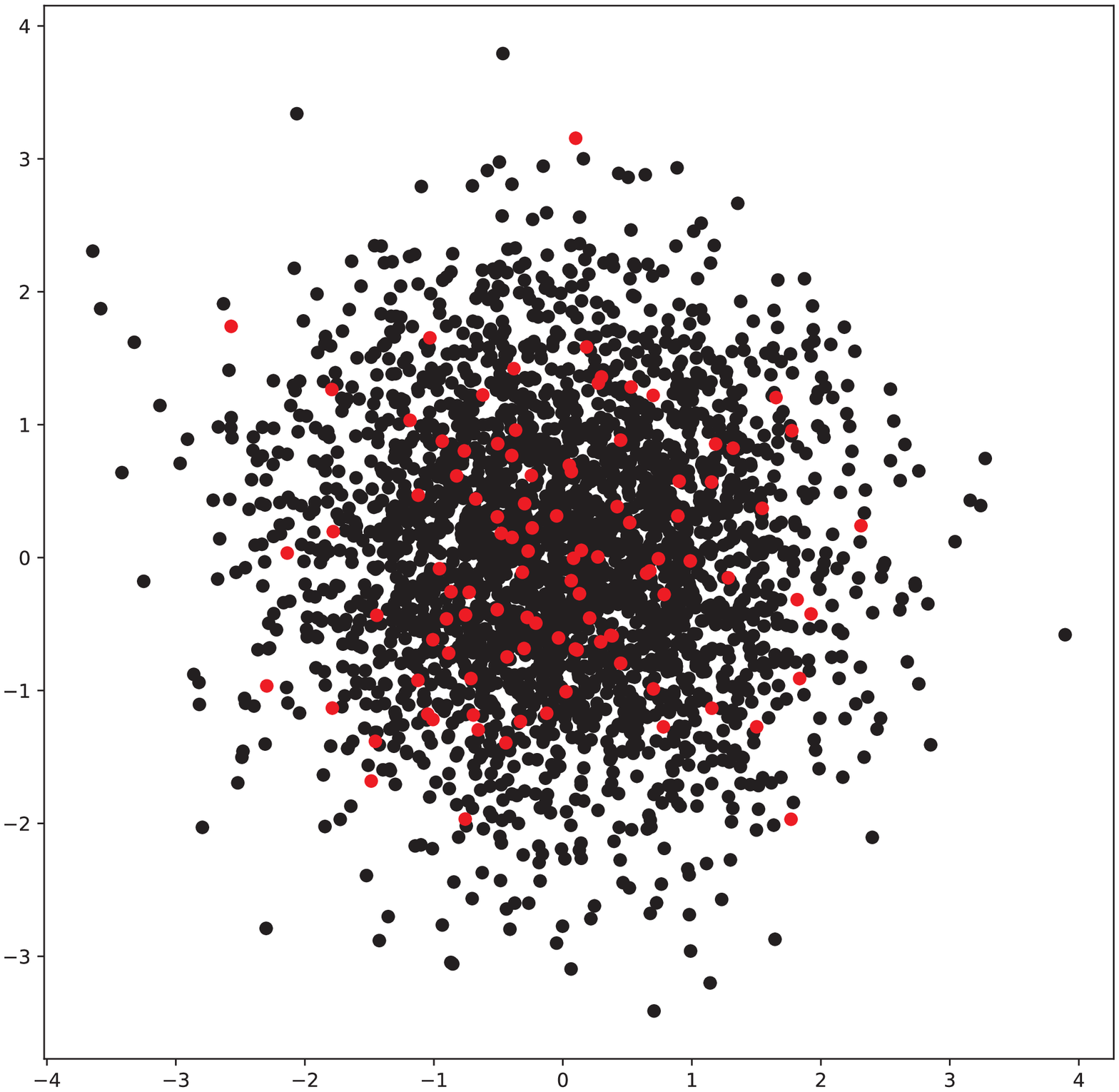}	}
    \subfigure[]{ \includegraphics[width=0.31\textwidth]{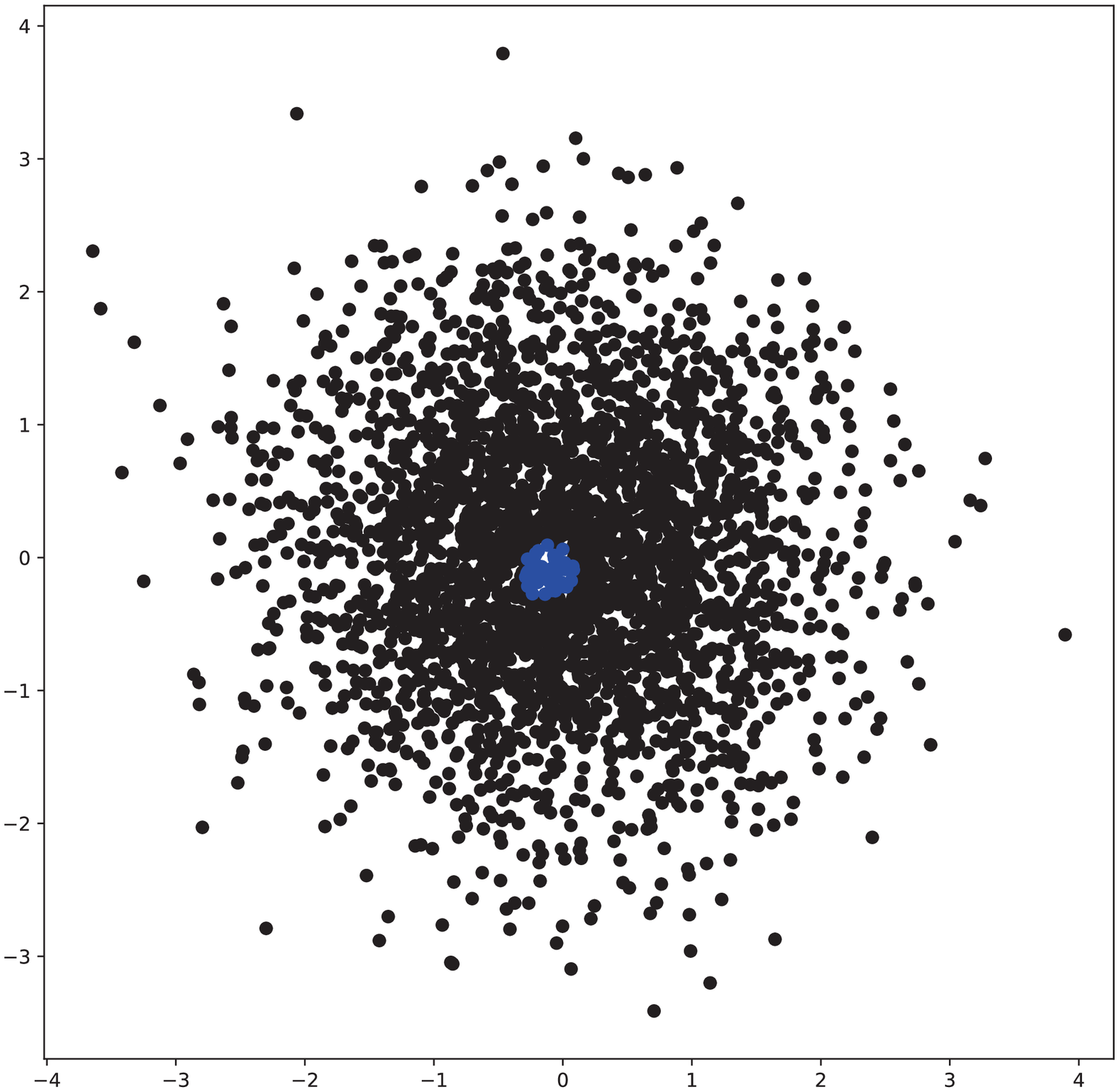}	}
    \caption{A toy example of proposed method. (a) 3,000 data points of standard normal distribution. (b) Randomly selected 100 (red) points. (c) Top 50 (blue) points corresponding to random points. }
\end{figure*}
\section{Background}
With respect to KMM problem, it aims to minimize the Maximum Mean Discrepancy (MMD) between the weighted distribution $ \alpha \left( x \right) $, the reference distribution $ { p }_{ r }\left( x \right)  $ and the matching distribution $ { p }_{ m }\left( x \right) $ in a Reproducing Kernel Hilbert Space (RKHS) $ \phi \left( x \right) :x\rightarrow h  $, such as
\begin{equation}\large
\begin{array}{ll}
	& { MMD }^{ 2 }\left( h ,\left( \alpha \left( x \right) ,{ p }_{ m }\left( x \right)  \right) ,{ p }_{ r }\left( x \right)  \right) \\ 
 & =  { \left\| { E }_{ x\sim { p }_{ m }\left( x \right)  }\left[ \alpha \left( x \right) \cdot \phi \left( x \right)  \right] -{ E }_{ x\sim { p }_{ r }\left( x \right)  }\left[ \phi \left( x \right)  \right]  \right\|  }^{ 2 }\\
 & = \arg \mathop {\min }\limits_\alpha  {\left\| {\frac{1}{{{n_{m}}}}\sum\limits_{i = 1}^{{n_{m}}} {\alpha \left( {{x_i}} \right)\phi \left( {{x_i}} \right)}  - \frac{1}{{{n_{r}}}}\sum\limits_{i = 1}^{{n_{r}}} {\phi \left( {{x_i}} \right)} } \right\|^2}\\
 & = \arg \mathop {\min }\limits_\alpha  \left[ {\frac{1}{{n_{m}^2}}\sum\limits_{i,j = 1}^{{n_{m}}} {{\alpha _i}k\left( {{x_i},{x_j}} \right){\alpha _j} - } } \right.\\
 & {\left. {\frac{2}{{{n_{m}}{n_{r}}}}\sum\limits_{i = 1}^{{n_{m}}} {\sum\limits_{j = 1}^{{n_{r}}} {{\alpha _i}k\left( {{x_i},{x_j}} \right)} }  + \frac{1}{{n_{r}^2}}\sum\limits_{i,j = 1}^{{n_{r}}} {k\left( {{x_i},{x_j}} \right)} } \right]}
\end{array}
\end{equation}
Here, $ k \left( x_i, x_j \right) $ indicates the kernel value calculated based on $ x_i $ and $ x_j $, and \emph{Gaussian} kernel is generally adopted.

By removing the constant item $ \frac { 1 }{ { n }_{ r }^{ 2 } } \sum _{ i,j=1 }^{ { n }_{ r } }{ k\left( { x }_{ i },{ x }_{ j } \right)  }  $, the objective can be redefined as
\begin{equation}\label{Eq-1}
	J\left( \alpha  \right) =arg\underset { \alpha  }{ min } \left[ \frac { 1 }{ 2 } { \alpha  }^{ T }{ K }_{ m, m }\alpha -\frac { { n }_{ m } }{ { n }_{ r } } \alpha { K }_{ m, r }e \right],
\end{equation}
where $ e $ denotes the unitary vector of all elements in one with suitable length. Obviously, this is a standard second-order optimization problem with convex optimization.
According to quadratic optimization, the ideal solution is calculated by setting objective gradient to be zero.
As a result, the ideal $ \alpha $ can be analytically obtained with a penalty item, e.g.,
\begin{equation}\large
 \alpha =\frac { { n }_{ m } }{ { n }_{ r } } { \left( { K }_{ m, m }+\lambda I \right)  }^{ -1 }{ K }_{ m, r }e,
\end{equation}
where $ I $ denotes the identity matrix, and $ \lambda $ is a constant with quite small value. Obviously, such results can be obtained by several direct approaches, e.g., quadratic optimization, least square regression. To conduct the problem, there have been many state-of-the-art methods proposed in the literature, and most of them refer to general quadratic optimization \cite{Miao15enKMM}\cite{Miao15LADR}\cite{Kanamori09LSDIE}.

On the other hand, there also exist some theoretical analysis on KMM under statistical learning conceptions \cite{Scholkopf01ES}\cite{Yu12KMM}\cite{Cortes10Bounds}, and several solutions have been proposed in the literature. The straight idea models the importance $ w \left( x \right) $ by the linear model in estimation of the Kullback-Leibler Importances \cite{Sugiyama07DIE}, and covariate shift adaptation is considered to be challenged in general \cite{Sugiyama07EGECS}\cite{Sugiyama09DIECSA}. In this work, the quadratic optimization framework is adopted, due to its convenient implementation and stable performance.

\section{Adaptive Matching of Kernel Means}
The basic idea of quadratic optimization of KMM is quite understandable, and employed widely in certain senses. Nevertheless, most existing methods depend on general solution to KMM objective, and adaptive characteristics have been deeply exploited in quite few works. 

\subsection{Kernel Mean Matching with Global Importance}
In order to make adaptive learning applicable, it is to bring self-taught learning to KMM. 
With respect to main idea of KMM, stable performance is desired to be conserved, as well as self-adaptive mechanism truly. The initial attempt to such topic can be referred to ensemble KMM \cite{Miao15enKMM}.
Actually, a natural consideration in KMM is to select the reference instances with great importance so that calculation cost can be reduced, which is referred as self-importance in general, i.e.,
\begin{equation}
	\widetilde {{w_i}} = \int_{r} {\phi \left( {x_i^{r}} \right)dx}  = \sum\limits_{j = 1}^{{n_{r}}} {k\left( {x_i^{r},x_j^{r}} \right)} 
\end{equation}
or equivalently,
\begin{equation}\label{Eq-2}
	\omega   { \left( {x_i^{r}} \right) } = \frac{{\int_{r} {\phi \left( {x_i^{r}} \right)dx} }}{{\sum\limits_{j = 1}^{{n_{r}}} {\widetilde {{w_j}}} }} = \frac{{\sum\limits_{j = 1}^{{n_{r}}} {k\left( {x_i^{r},x_j^{r}} \right)} }}{{\sum\limits_{j = 1}^{{n_{r}}} {\widetilde {{w_j}}} }}.
\end{equation}
As a consequence, the importance of each instance can be estimated accordingly, and further mean matching is proceeded with reduced densities. For convenience, such adaptive KMM method is named as global KMM (gloKMM) for short in the context, as global importance of instances are considered for selection. The whole procedure of global KMM algorithm is given in Algorithm 1.
\begin{algorithm}
\caption{The global KMM (gloKMM) algorithm}
\KwIn{ Given matching instances  $ { x }_{ i }^{ m }~ \left( i=1,2,\cdots ,{ n }_{ m } \right) $, reference set $ { x }_{ i }^{ r }~ \left( i=1,2,\cdots ,{ n }_{r } \right) $, desired number of reference instances $ n_h $ with highest importance. }
\KwOut{The estimated importance $ w \left( x \right) $.}
1. Calculate the importance of each reference instance as done in (\ref{Eq-2}), and select the $ n_h $ instances with highest importance.	\\
2. Calculate the kernels $ K_{m, m} $ and $ K_{m, h} $ with selected matching and reference instances.	\\
3. Solve the KMM problem in (\ref{Eq-1}) and obtain the optimal coefficients $ \alpha $.		\\
4. Calculate estimated importance of instances by $ w\left( x \right) $.
\end{algorithm}

Obviously, such idea focuses on the global importance of reference instances $ x^r $, and most important instances are referred.
In addition, it is tractable to adopt generalized inverse for calculation of $ \alpha $, while affordable training instances are usually involved in KMM, e.g., 
\begin{equation}
	{ K }_{ m,m }^{ + }=V{ S }^{ + }{ U }^{ T },
\end{equation}
where $ U $ and $ V $ are the left and right orthogonal matrices obtained from singular value decomposition of $ K_{m, m} $  as $ K_{m, m} = U S V^T $, and $ S^+ $ is the diagonal matrix with inverse of eigenvalues putting along the diagonal.
Nevertheless, the iterative optimization may be still necessary for large-scale data set, owing to the fact that inverse of large matrix is hardly to be calculated directly. Furthermore, it is clear that complexity of estimation of importance mainly depends on amount of reference data.

After obtaining $ \alpha $, it is definitely to calculate the importance of instances from matching and reference sets. Without loss of generality, the \emph{Gaussian} kernel model centered at the test points is adopted, i.e.,
\begin{equation}
	\widehat { w } \left( x \right) =\sum _{ i=1 }^{ { n }_{ m } }{ { \alpha  }_{ i }k_{ ga }\left( x,{ x }_{ i }^{ m } \right)  }.
\end{equation}
Accordingly, the differential importance of different sets can be disclosed, and novelties of knowledge contents are able to be discovered.

\subsection{Adaptive Matching of Kernel Means}
Though global KMM method is able to achieve tenable outlets with the importance of top instances, there still exist limitations of adaptive matching. More specifically, it is found the performance of global KMM has been quite related with referred data set, and it fails to give acceptable results with changeable instances.
And the calculation of such global KMM method is the importance estimation of instance selection is based on all reference instances. As a consequence, it could be hardly to keep the efficiency if large amount of data are desired, and scalable learning is unable to be reached. It leads to the motivation of this work in designing of a flexible development to conquer the challenge of matching performance with adaptive learning idea. In terms of this, an improved KMM solution is devised to reduce the possible complexity of global KMM, while both scalable extension and the stability of algorithmic performance are reached.

The basic idea of proposed method is to reduce the calculation complexity of importance estimation, and especially, the complete estimation round of whole instances. Thereafter, it is to select a subset of reference data for estimation of importance, and it is verified the estimated importance results in acceptable ranking of reference data. 
Furthermore, the random selection mechanism is preferred in AMKM accordingly. 
As a consequence, the modified estimation of instance importance is defined as
\begin{equation}\label{Eq-3}
  \omega  { \left( {x_i^{r}} \right) } = \frac{{\int_{{n_{s}}} {\phi \left( {x_i^{r}} \right)dx} }}{{\sum\limits_{j = 1}^{{n_{s}}} {\widetilde {{w_j}}} }} = \frac{{\sum\limits_{j = 1}^{{n_{s}}} {k\left( {x_i^{r}, x_j^{r}} \right)} }}{{\sum\limits_{j = 1}^{{n_{s}}} {\widetilde {{w_j}}} }},
\end{equation}
where $ n_{s} $ denotes the amount of selected instances from data set. Nevertheless, it is hardly to ensure the importance of instances with randomly selected data in loss of acceptable matching. To address this problem, a re-choosing mechanism is designed to pick up the instances with highest importance. More specifically, a selective importance estimation is performed based on randomly selected instances as reference, and the top important instances are selected as matching subset, which is referred as a \emph{refinement} stage. A toy example of proposed method is given in Fig. 1. As illustrated, the refinement stage is able to get the approximately centered instances corresponding to random references, which contributes to mean matching of global distribution of data. Nevertheless, the obtained matching results generally rely on unaccurate means, and further calibration may be necessary for improvement. To address this limitation, the randomly selective matching is repeated several times, and a fusion stage is adopted to learn the ideal matching.

Suppose that, there are $ t $ approximately matching results, i.g., $ \begin{array}{*{20}{c}}
{{{\rm M}_i} = [{\alpha _{i,1}},{\alpha _{i,2}}, \cdots ,{\alpha _{i,{n_s}}}],}&{i = 1,2, \cdots ,t}
\end{array} $, which are obtained by repeated adaptive learning. Then, the final matching is calculated via a fusion of different $ \rm M_i $ with optimization of objective of KMM. Without loss of generality, it can be defined as a combination of different coefficients corresponding to matching objective, e.g.,
 \begin{equation}
 \begin{array}{ll}
 J\left( \beta  \right) & = \arg \mathop {\min }\limits_{{\beta _i}} \sum\limits_{i = 1}^t {\sum\limits_{j = 1}^{{n_s}} {\left( {\frac{1}{2}\gamma _{_{i,j}}^T{K_{m,m}}{\gamma _{i,j}} - \frac{{{n_m}}}{{{n_r}}}{\gamma _{i,j}}{K_{m,r}}e} \right)} } \\
{with}&{{\gamma _{i,j}} = {\alpha _{i,j}}{\beta _i}}
\end{array}
 \end{equation}
Obviously, the ideal $ \beta_i $ can be calculated by optimizing such objective via a standard quadratic programming (QP) with fixed $ \alpha_{i, j} $ \cite{Boyd04CO}\cite{Bertsekas15CO}\cite{Bubeck15CO}. As a traditional consideration, the constraints of such QP can be referred to certain equivalent conditions of $ \beta_i $ as well as the lower or upper bounding. Nevertheless, it is practically demonstrated that valid solutions are hardly to be attained with full constraints, especially for summation of $ \beta_i $. As a consequence, the relaxed constraint conditions are adopted to restrict $ \beta_i $ to be values larger than zero only, and the objective can be further defined as
\begin{equation}\label{Eq-3}
\begin{array}{ll}
J\left( \beta  \right) & = \arg \mathop {\min }\limits_{{\beta _i}} \sum\limits_{i = 1}^t {\sum\limits_{j = 1}^{{n_s}} {\left( {\frac{1}{2}\gamma _{_{i,j}}^T{K_{m,m}}{\gamma _{i,j}} - \frac{{{n_m}}}{{{n_r}}}{\gamma _{i,j}}{K_{m,r}}e} \right)} }	\\
{with}&{{\gamma _{i,j}} = {\alpha _{i,j}}{\beta _i}}	\\
{s.t.}&{{\beta _i} \ge 0}		\\
\end{array}
\end{equation}
Thereafter, the ideal matching can be adaptively learned by solving such QP with pre-learned several matching in complexity $ O \left( t \right) $. The whole procedure of adaptive KMM is given in Algorithm 2.


Thereafter, the calculation complexity can be significantly improved, as both of global KMM and AMKM learn the mean matching with subsets of reference data. 
Nevertheless, the main difference between them relies on the selection of important reference instances. The global KMM selects the important instances with calculation of importance of whole data, while AMKM adopts selective important instances with a refinement step and repeated learning is employed to enhance the performance of random selection. Furthermore, the fusion of different random matching is optimized via optimizing a QP objective, which alleviates the negative influence of matching of approximately centered kernel means. As a consequence, the calculation of importance of whole instances can be avoided in AMKM owing to adaptively selective matching of important reference instances, and much efficiency can be preserved. To highlight the advantages of the proposed method, it is explicitly summarized as below.
\begin{itemize}
\item Due to selective matching, it is unnecessary to involve the full kernel matrix into learning procedure, which is quite different from standard KMM.
\item To improve the complexity of selection of important reference data, the refined selection is adopted to find the instances with highest importances. And calculation of importance of whole instances can be avoided for selection.
\item The fusion of matching results associated with different selective references can be adaptively learned via solving a QP optimization, which can be calculated in efficiency.  
\end{itemize}
\begin{algorithm}
\caption{The AMKM algorithm}
\KwIn{ Given matching instances  $ { x }_{ i }^{ m }~ \left( i=1,2,\cdots ,{ n }_{ m } \right) $, reference set $ { x }_{ i }^{ r }~ \left( i=1,2,\cdots ,{ n }_{ r } \right) $, number of repeation $ t $, number of randomly selected instances $ n $, desired number of important instances $ n_s $ for matching. }
\KwOut{The estimated importance $ w \left( x \right) $. }
\While{The desired repetition $ t $ has never reached}
{
1. Randomly select $ n $ instances from $ { x }_{ i }^{ r } $.	\\
2. Choose the most important $ n_s $ instances from reference data associated with the previously selected $ n $ instances.		\\
3. Follow the steps 2-3 in Algorithm 1.		\\
}
4. Calculate the fusion coefficients by solving the QP defined in (\ref{Eq-3}).		\\
5. Calculate estimated importance of samples $ w\left( x \right) $.
\end{algorithm}

\subsection{Discussion}
Though similar partition and fusion stages are involved, it is worthwhile to differentiate the proposed method from other methods. Both AMKM and ensemble KMM \cite{Miao15enKMM} refer to selection of reference instances, however, ensemble KMM relies on partition of reference set and the complete set is still absorbed. Contrarily, AMKM performs the selection with a separate refinement stage. More specifically, AMKM firstly randomly select a few instances, and then takes the reference subset with the highest importance corresponding to those pre-selected instances. Compared with ensemble KMM, AMKM randomly selects the subset of reference data with no explicit rule, and the volume of referred data can be changed conveniently. 
Furthermore, the fusion stage is employed to combine the matching results of different repeated procedures. Different from existing methods, the ideal fusion is learned by re-optimizing the original KMM objective, which is calculated via solving a QP problem. In the literature, the similar idea was proposed to find the optimized KMM with second-order optimization \cite{Gong14LK} straightforward. Nevertheless, AMKM only adopts QP to learn the ideal fusion of repeated learning, and KMM is efficiently performed associated with different subsets of reference data, which is calculated indirectly.

The essential idea of AMKM is to select the most important instances with respect to a subset of random instances, but specific calculations are devised for improvement of efficiency. There are several ways to disclose the theoretical bases of such method. For example, the information theory can be adopted to make a discussion, and other related fields can be inferred accordingly. Without loss of generality, the measure of selection of instances is identical with information potentials \cite{Erdogmus02IP}, which is defined as  
\begin{equation}
 V\left( {{x^r}} \right) = \frac{1}{{n_s^2}}\sum\limits_{i = 1}^{{n_s}} {\sum\limits_{j = 1}^{{n_s}} {G\left( {x_i^r - x_j^r,2{\sigma ^2}} \right)} }. 
\end{equation}
Here, $ {G\left( {x_i^r - x_j^r,2{\sigma ^2}} \right)} $ denotes the symmetric Gaussian kernel, $ n_s $ denotes the amount of selected instances. Furthermore, the Renyi quadratic entropy can be succinctly written as
\begin{equation}
 H\left( x \right) =  - \int_{{x^r}} {\log {p^2}\left( {{x^r}} \right)dx = }  - \log V\left( {{x^r}} \right).
\end{equation}
Obviously, the selected important instances can be explained as the ones corresponding to the maximum information potentials of the pre-selected random instances, and the minimum disorder of data as well. And it is known that, the centralized instances hold the most informative patterns in general. This is identical with depiction of Gaussian kernels and the intrinsic distribution of data.

\begin{figure*}
    \centering
    \subfigure[]{ \includegraphics[width=0.31\textwidth]{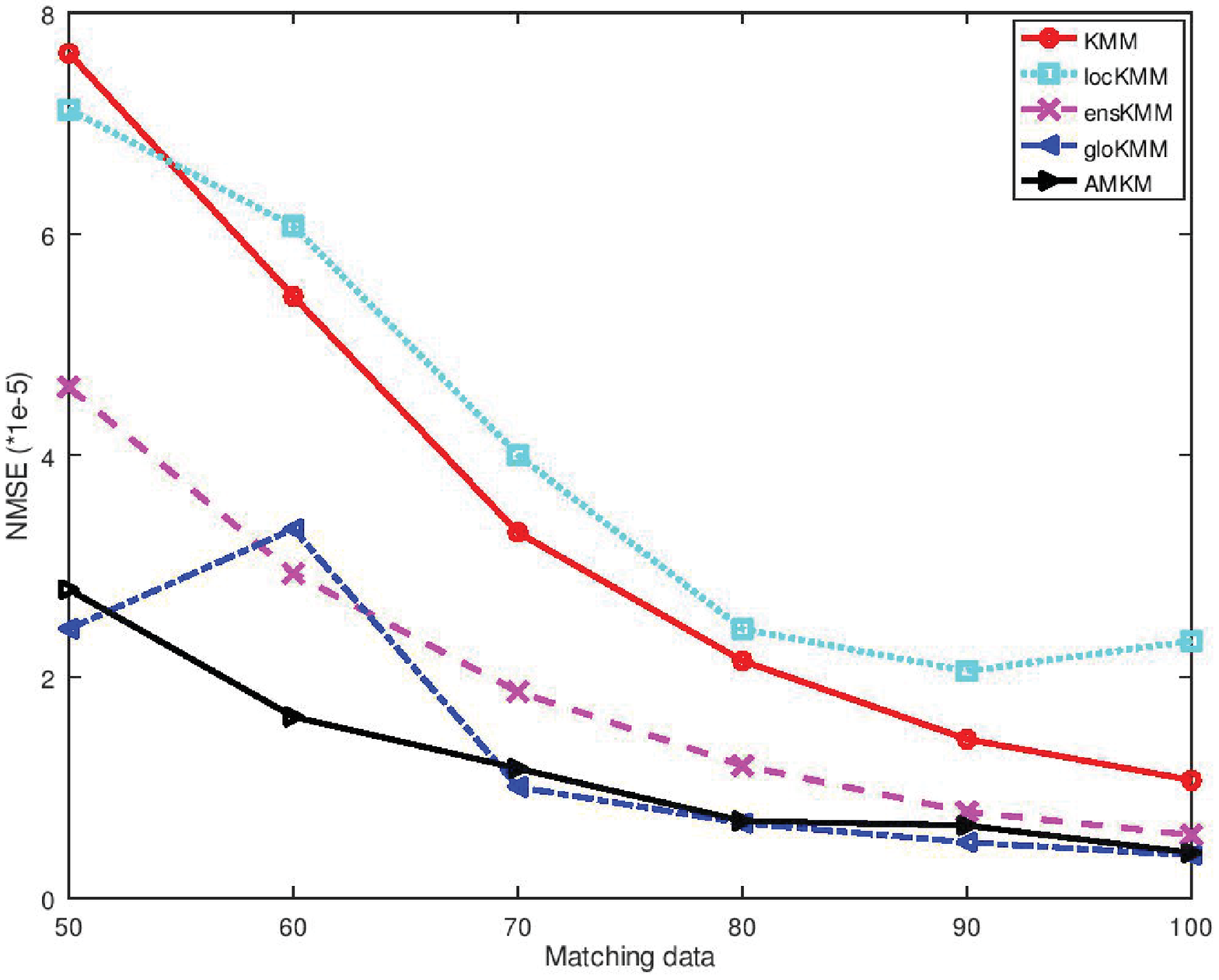}		}
    \subfigure[]{ \includegraphics[width=0.31\textwidth]{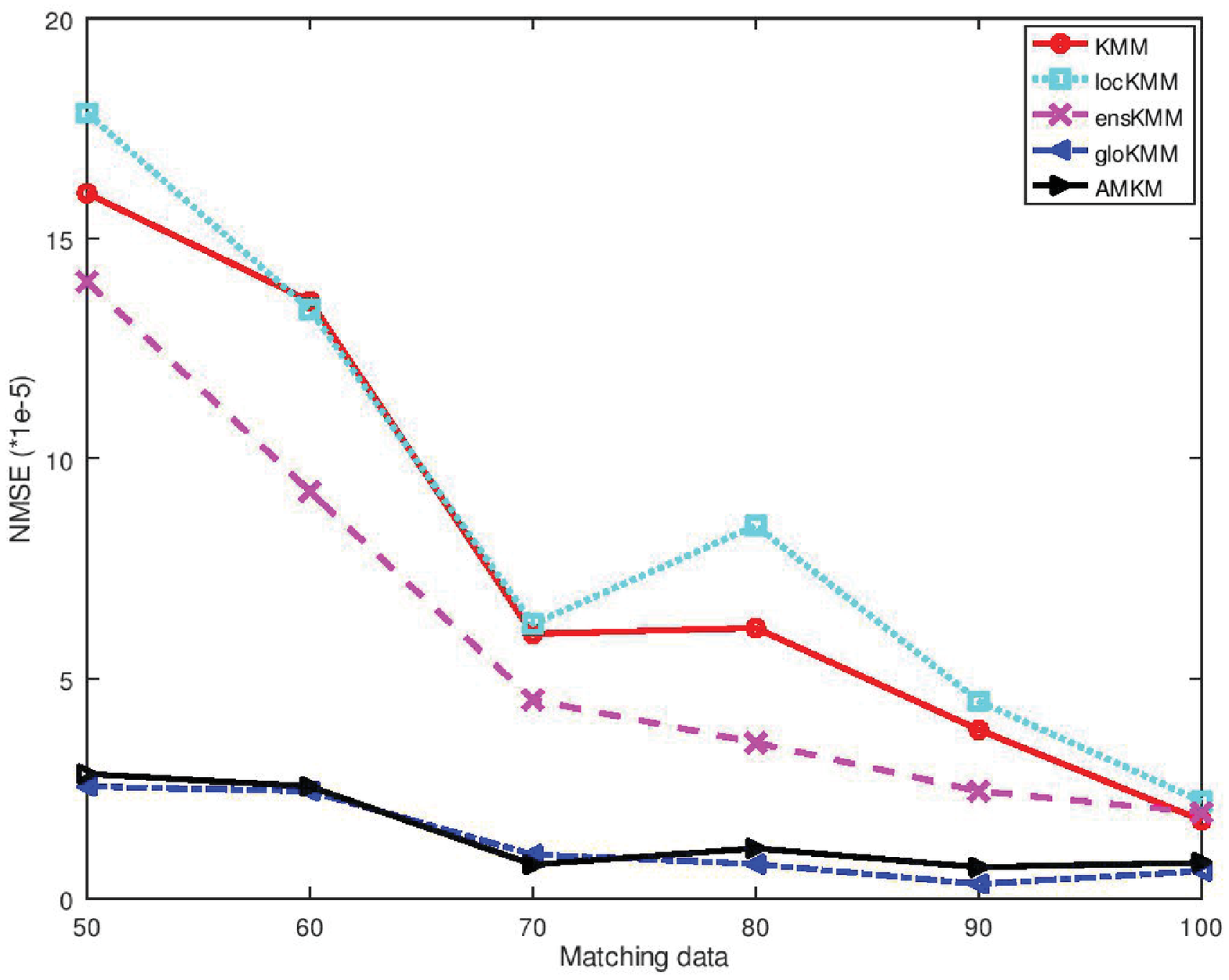}	}
    \subfigure[]{ \includegraphics[width=0.31\textwidth]{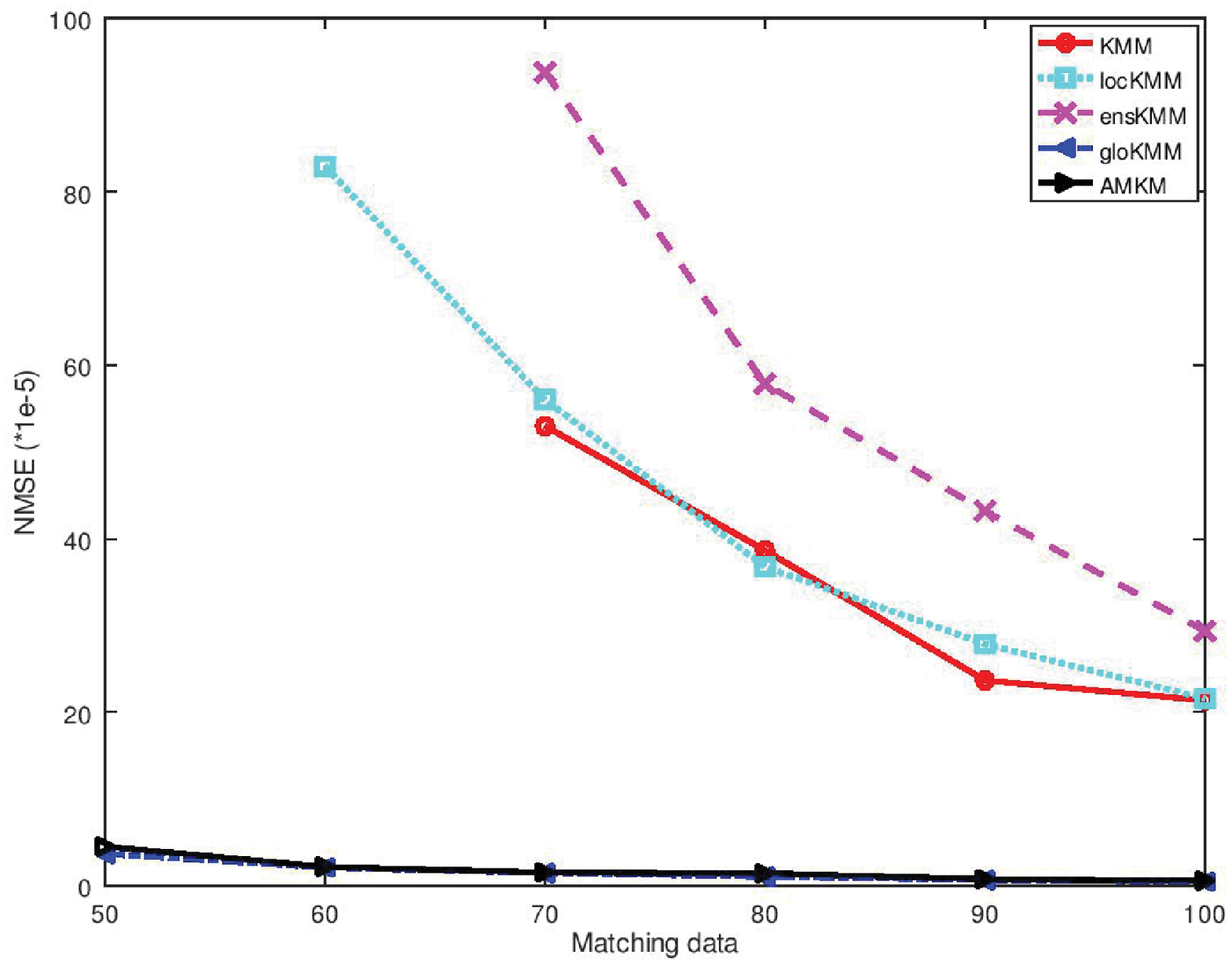}		}
    \subfigure[]{ \includegraphics[width=0.31\textwidth]{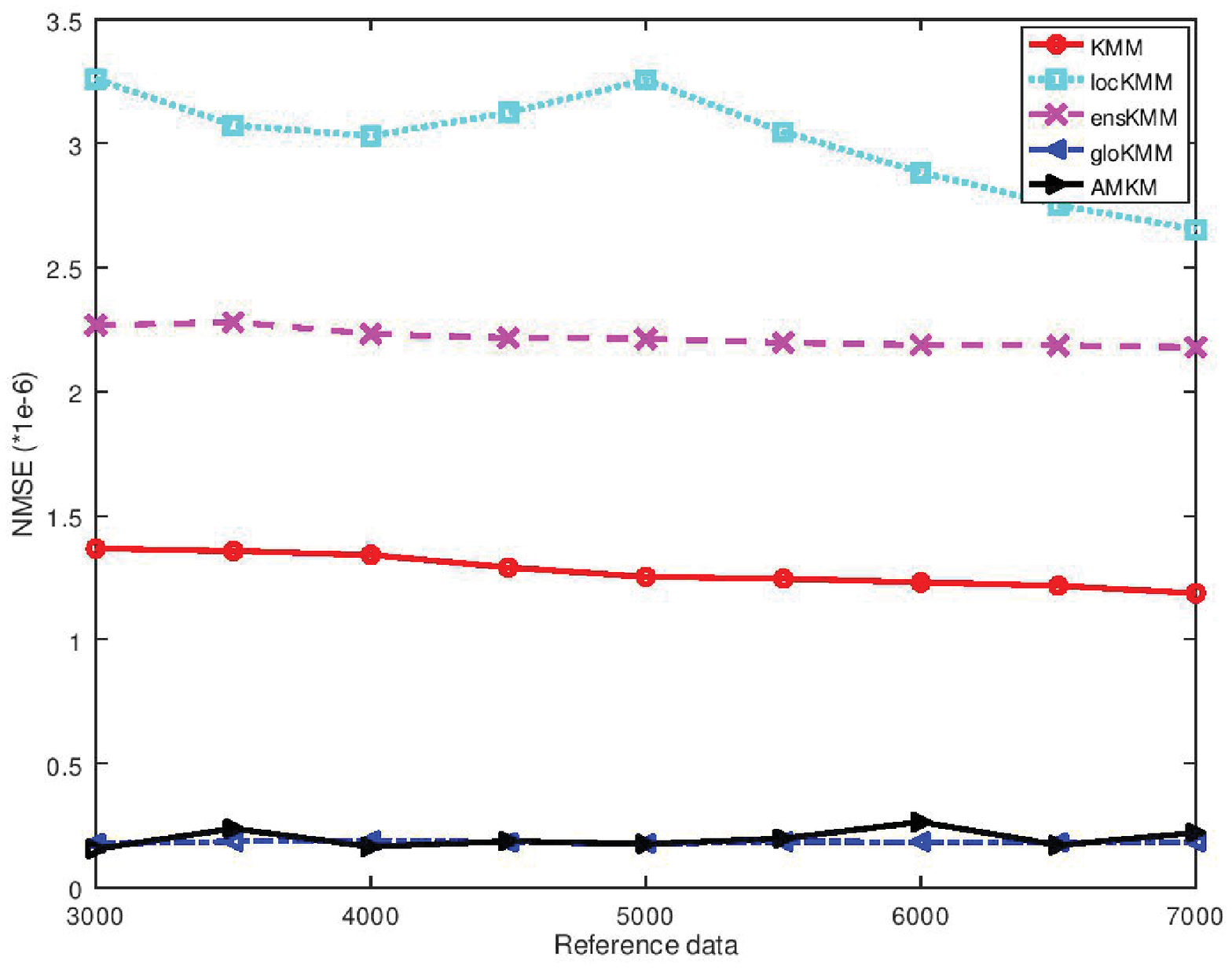}		}
    \subfigure[]{ \includegraphics[width=0.31\textwidth]{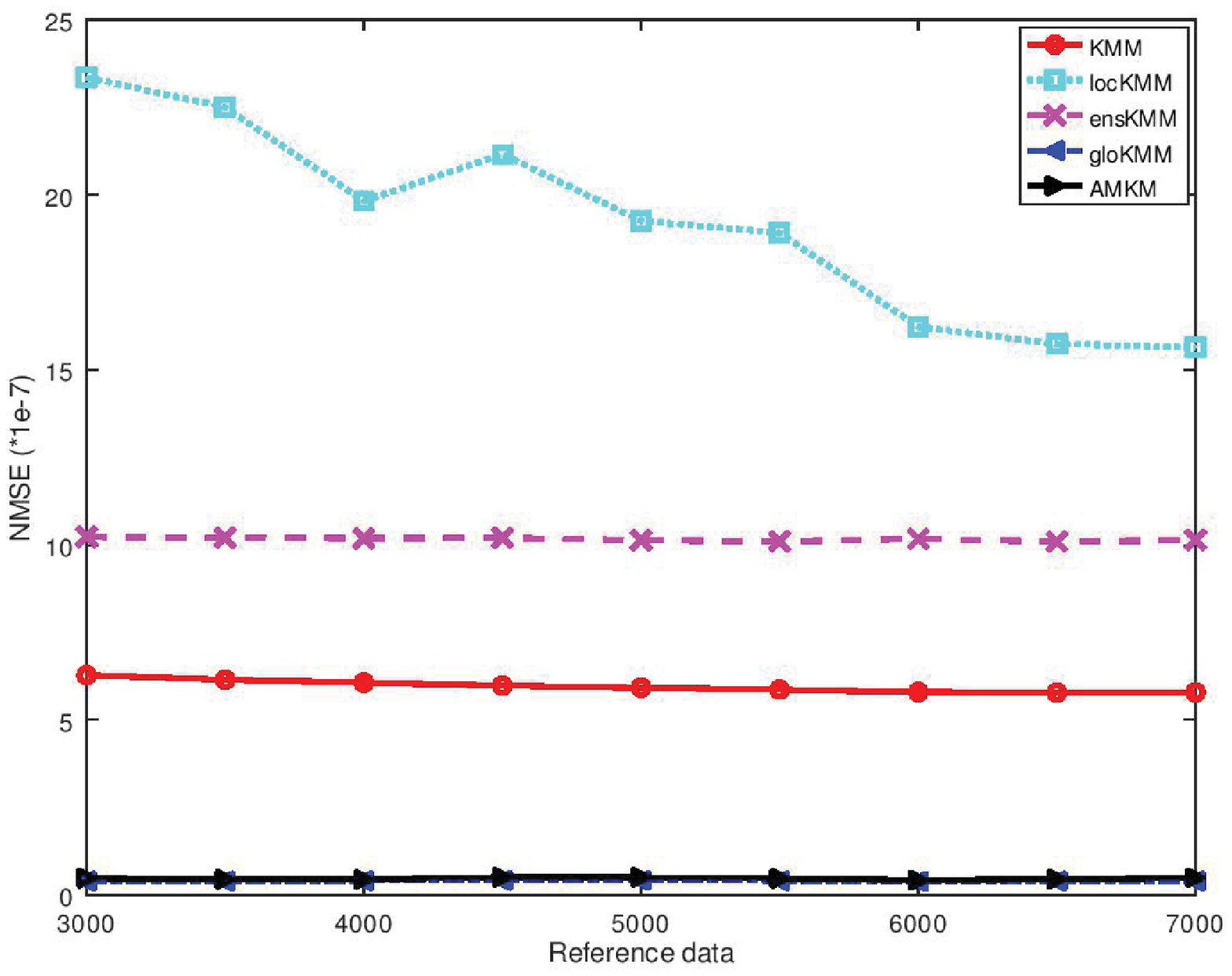}		}
    \subfigure[]{ \includegraphics[width=0.31\textwidth]{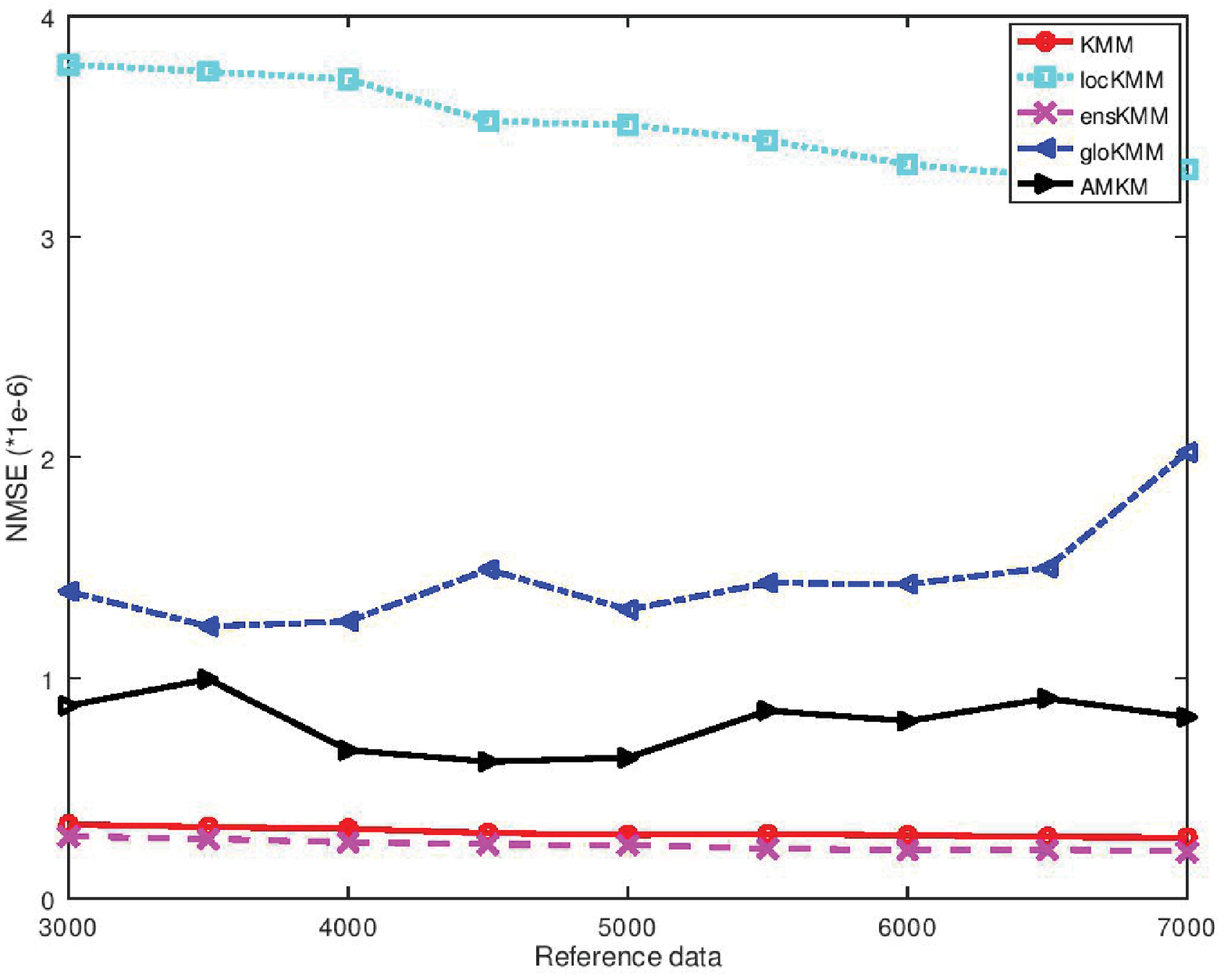}		}
    \caption{The experimental results of different KMM methods on various data sets. (a)-(c): The obtained NMSE on Monks, Ionosphere, and Climate data sets with different sizes of matching data. (d)-(f): The obtained NMSE on Forest, Letter and CIFAR data sets with different sizes of reference data.}
\end{figure*}
\section{Experiments}
In this section, the efficiency of proposed method are evaluated with several state-of-the-art methods, 
i.g., standard KMM \cite{Kanamori09LSDIE}, locally KMM (locKMM) \cite{Miao15LADR}, ensemble KMM (ensKMM) \cite{Miao15enKMM}, global KMM (gloKMM). Six artificial data sets \cite{Dua:2019}\cite{doornik1994practical}, namely three data sets with normal sizes and another three large data sets, are used in the experiments.
Some details of these data sets are given in Tab. I. For CIFAR-100 data set, the deep features of samples are learned, and then the feature data of first 20 categories are selected to form a subset, which consists of 500 samples in each category.
\begin{table}[h]
	\centering
	\caption{The details of different data sets}
	\label{tab:Margin_settings}
	\begin{tabular}{|l||c|c|}\hline
		Data Sets & Samples & Dimensionality	\\
		\hline \hline
		Monks  &  1,711  &  6 	\\
		Ionosphere  &  351  &  34		\\
		Climate  &  540  &  18		\\
		Forest  &  581,012  & 	54	\\
		Letter  &  20,000 &	16	\\
		CIFAR  &  10,000  & 	255	\\
		\hline
	\end{tabular}
\end{table}
The performance of different estimation methods are evaluated with the \emph{Normalized Mean Squared Error} (NMSE) \cite{Sugiyama07DIE}.
In addition, five neighbors are involved to calculate the locally adaptive kernels in local KMM, and five random partitions are applied in ensemble KMM. For global KMM, the top 100 important reference data are adopted, and similarly, 100 instances with the highest importances are selected with random 50 samples for calculation in adaptive KMM. All experiments of each algorithm is repeated \emph{five} times with different input instances\textsuperscript{1} \footnote{\textsuperscript{1} All experiments are performed on a Computer with hardware of Intel i5 2.8 GHz CPU and 16 GB Memory.}, and then the average of outputs are recorded as results.

To disclose the algorithmic efficiency of different algorithms, the reference and matching data sets are set to be fixed sizes alternatively while the size of the other data is changed during the first experiment. In detail, a fixed size of reference data is set to be 500, 250, 400 with random selection for the first three data sets. And different sizes of instances are selected to be the matching data, which are changed in range from 50 to 100. The experimental results are shown in Fig. 2 (a)-(c), while some results on Climate data set with the order of magnitude of $ 10^{-2} $ are ignored to make the illustration clear. As the matching sizes are increased with 50 samples, the gloKMM and AMKM are able to give the best results. Furthermore, the most stable performance are presented by gloKMM and AMKM, especially for matching of Ionosphere and Climate data sets. In most cases, the standard KMM and locally KMM can reach the close results, and better performance is given by KMM with large matching data. For Monks and Ionosphere data sets, ensKMM is able to give the moderate results between KMM and gloKMM algorithms. But worst results are obtained from ensKMM on Climate data. On the contrary, the performance of different methods with various sizes of reference data are evaluated on another three large data sets. Among instances of Forest, Letter and CIFAR data sets, respective 500 instances are randomly selected to be matching data, while instances in the range from 3,000 to 7,000 are selected to be the reference data during each execution. The experimental results are shown in Fig. 2 (d)-(f). Similarly, gloKMM and AMKM are able to get the best results, and ideal performance are obtained on Forest and Letter data sets, followed by standard KMM and ensKMM methods. Compared with experiments in different sizes of matching data, the obtained experimental results of most methods are quite stable. In other words, the matching results of different algorithms are insensitive to changing of reference sizes. Nevertheless, the best results on CIFAR data set are obtained by KMM and ensKMM methods. Though the results of gloKMM and AMKM are still close to each other, the second rank of results are obtained. In addition, it is noticeable that the differences of methods are quite small, which are bounded by the tiny order of magnitude.

\begin{figure*}
    \centering
    \subfigure[]{ \includegraphics[width=0.31\textwidth]{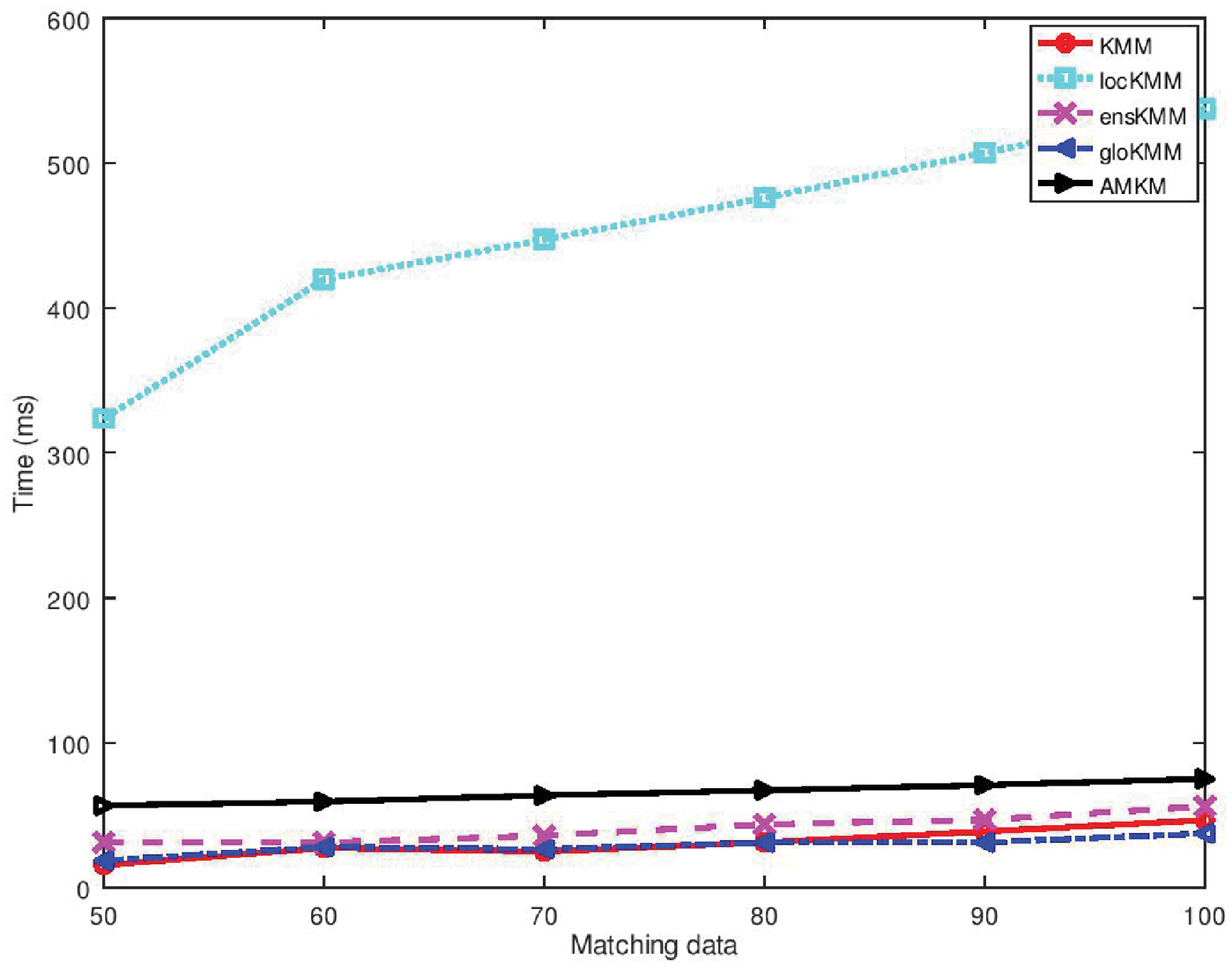}		}
    \subfigure[]{ \includegraphics[width=0.31\textwidth]{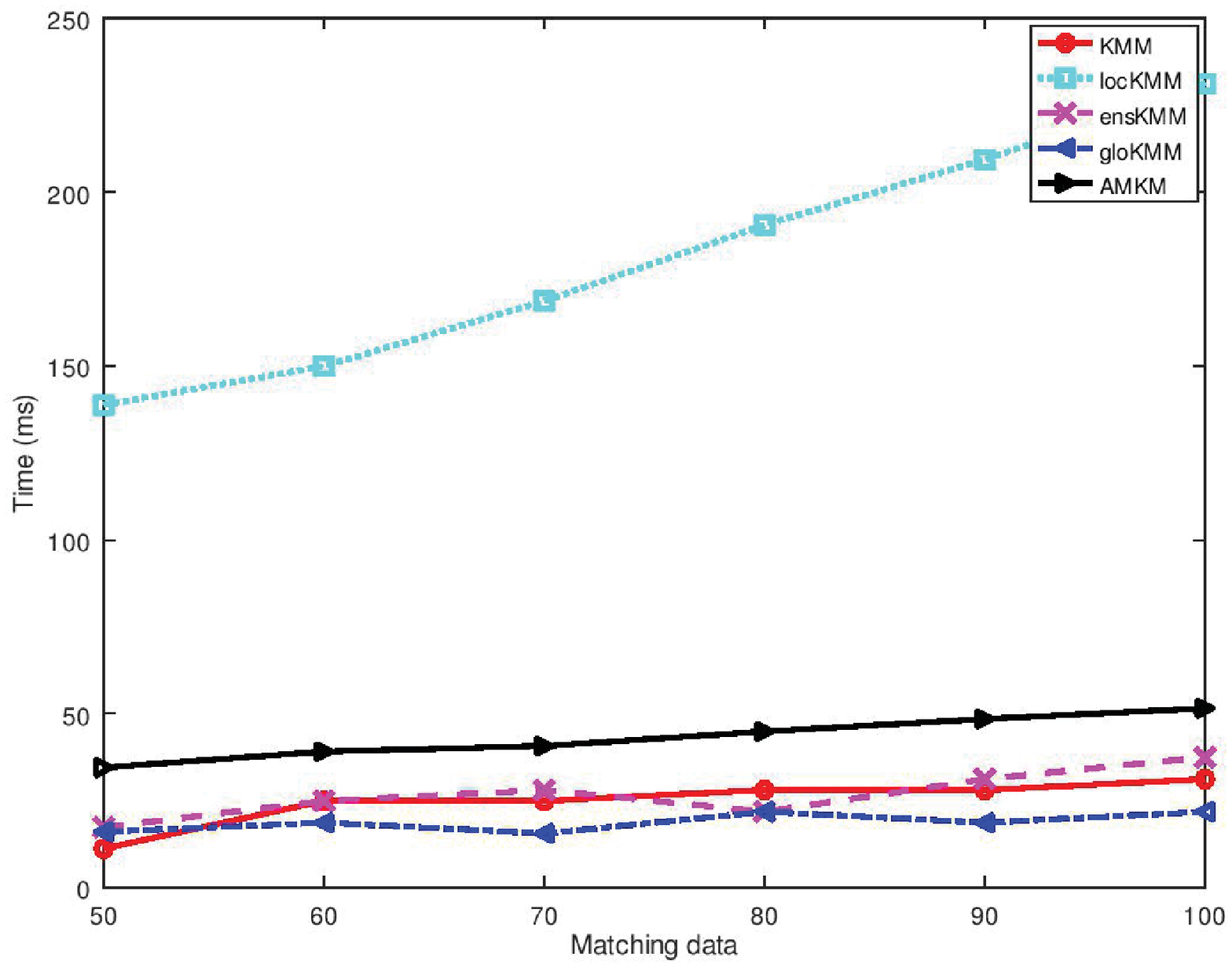}	}
    \subfigure[]{ \includegraphics[width=0.31\textwidth]{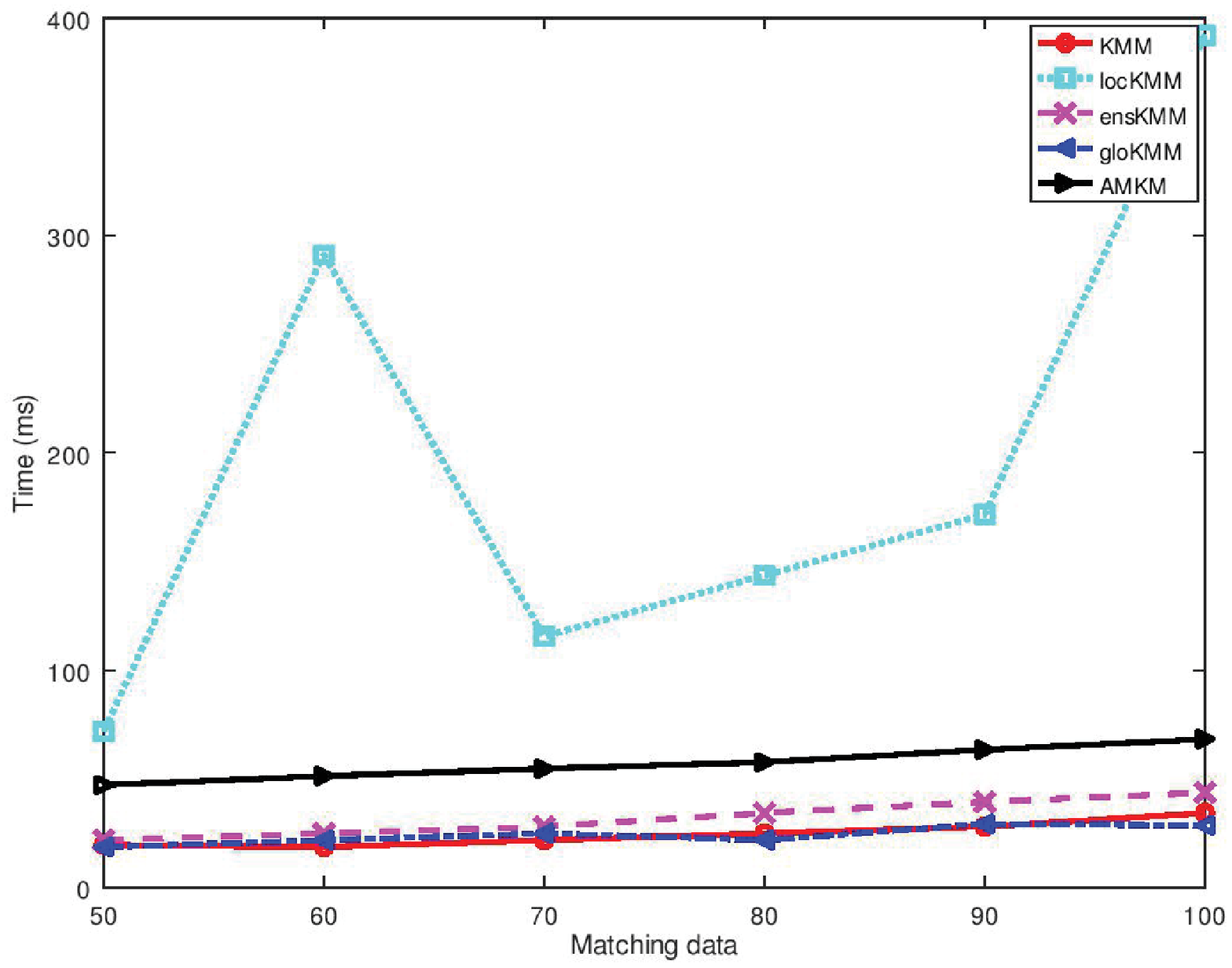}		}
    \subfigure[]{ \includegraphics[width=0.31\textwidth]{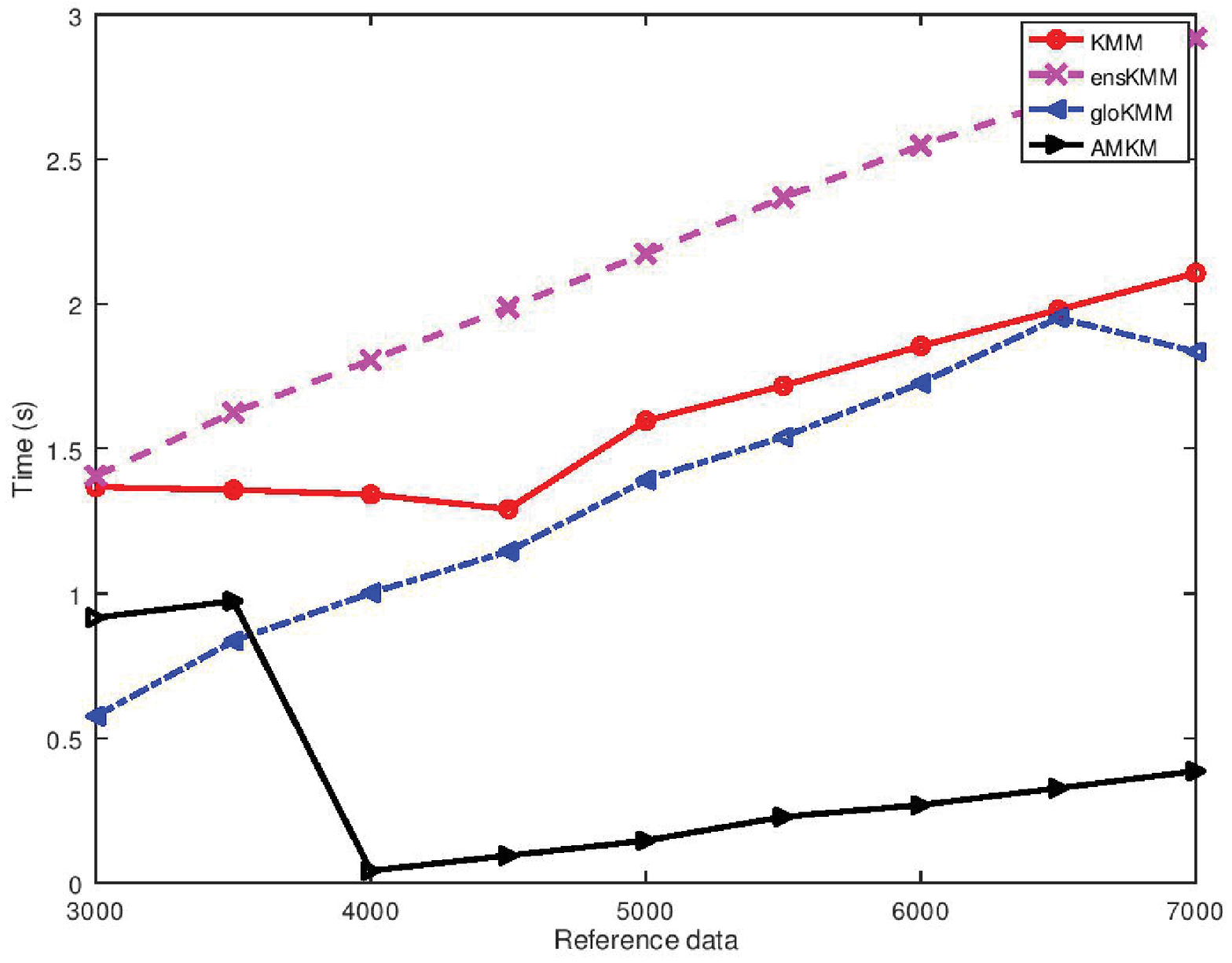}		}
    \subfigure[]{ \includegraphics[width=0.31\textwidth]{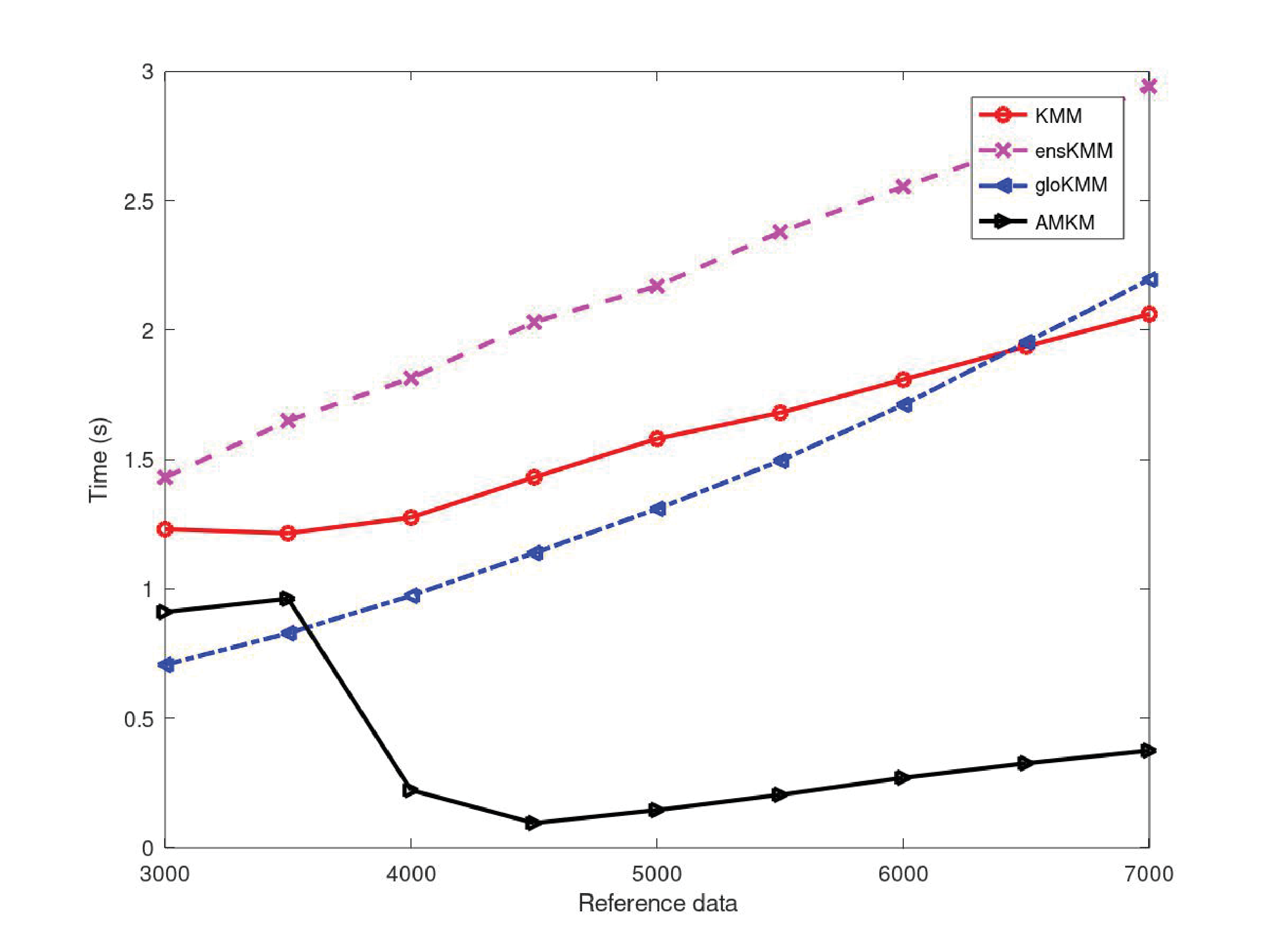}		}
    \subfigure[]{ \includegraphics[width=0.31\textwidth]{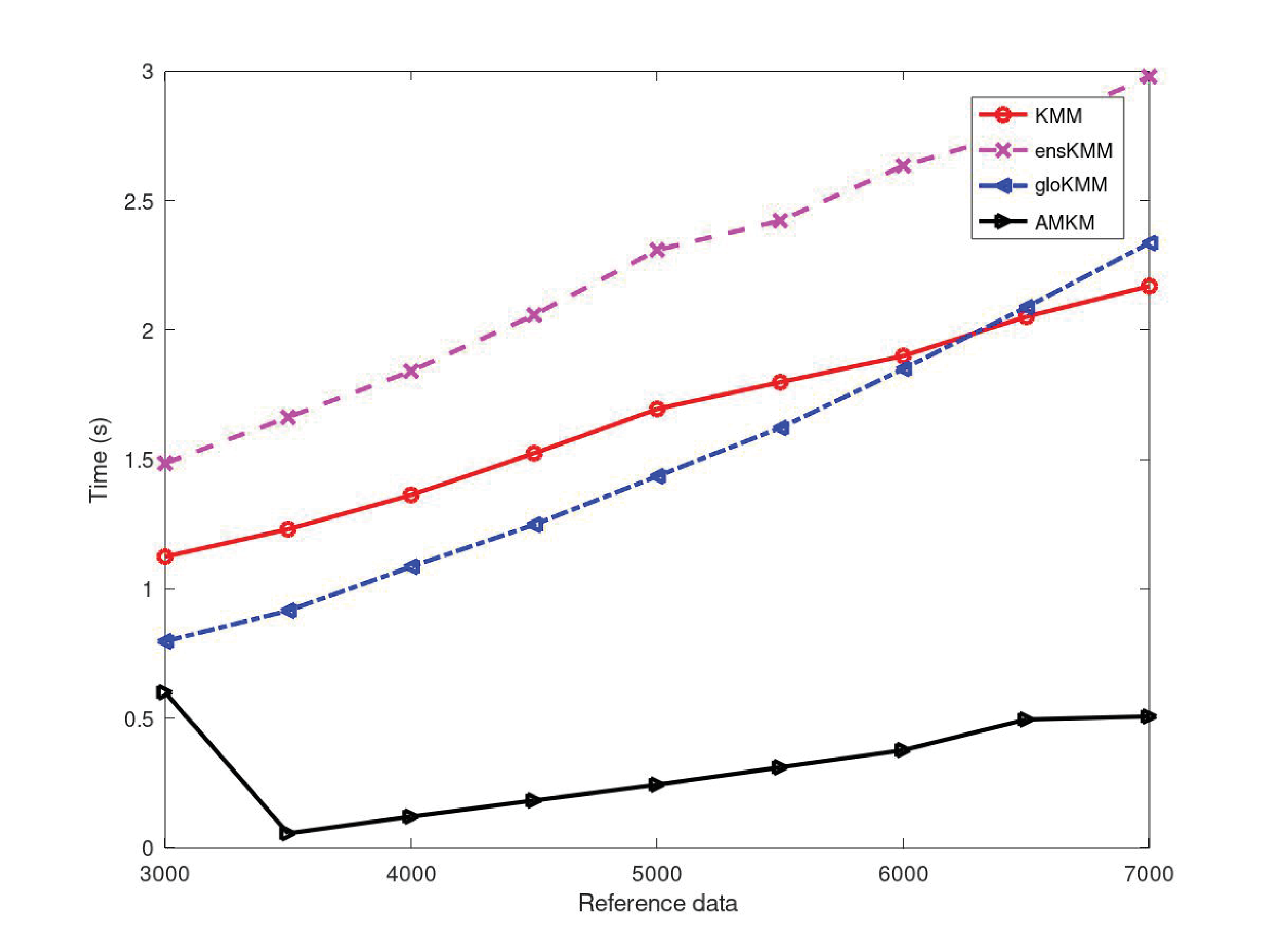}		}
    \caption{The time complexities of different KMM methods on various data sets. (a)-(c): The time complexities (milliseconds) of different algorithms on Monks, Ionosphere, and Climate data sets with different sizes of matching data. (d)-(f): The time complexities (seconds) of different algorithms on Forest, Letter and CIFAR data sets with different sizes of reference data.}
\end{figure*}
On the other hand, the time complexities of different methods during experiments are recorded, which are shown in Fig. 3. Similarly, the average cost times of \emph{five} repetitions are recorded as results for each algorithm. According to the experimental results on data sets with normal sizes, gloKMM is promised to give the most efficient performance, followed by KMM and ensKMM. And their costs are still close to each other with different sizes of matching data. Nevertheless, the calculation complexity of AMKM is inferior compared with those methods. This is due to the fact that limited instances are involved in experiments, and consequently, the calculation advantage of AMKM cannot be put to good use. For the experimental results on large data sets, the time complexity of locKMM takes ten more seconds, which are far from other methods for discussion. As the sizes of reference data are increased, the time complexities of different algorithms are augmented accordingly. Nevertheless, the costs of gloKMM and AMKM are superior to other methods. And AMKM is able to give more efficiency as reference data is enlarged. With the multi-folds learning, ensKMM presents the moderate efficiency compared with other methods but still superior to locKMM.

In the second experiment, the scalable learning ability of proposed methods are evaluated. For Forest and Letter data sets, 500 and 3,000 instances are respectively selected to be matching data and reference data. Then, reference data are appended with another 500 instances each time for batch matching. Similarly, \emph{five} repetitions are performed and average results are recorded. Furthermore, ensKMM is referred with scalable setting, and the obtained results are given in Fig. 4. In terms of experimental results, stable matching can be obtained for all methods. With increasing reference instances, AMKM and gloKMM are able to achieve the quite similar matching results, and outperforms the ones obtained by ensKMM. And it is worthwhile to note that, small resulting differences of methods are obtained, which are depicted by the tiny order of magnitude. In other words, AMKM can achieve approximate matching performance to gloKMM while efficiency is preserved for KMM.

\begin{figure*}
    \centering
    \subfigure[]{ \includegraphics[width=0.48\textwidth]{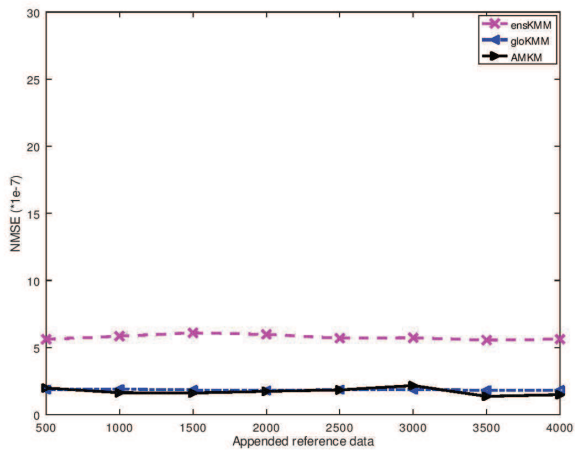}	}
    \subfigure[]{ \includegraphics[width=0.48\textwidth]{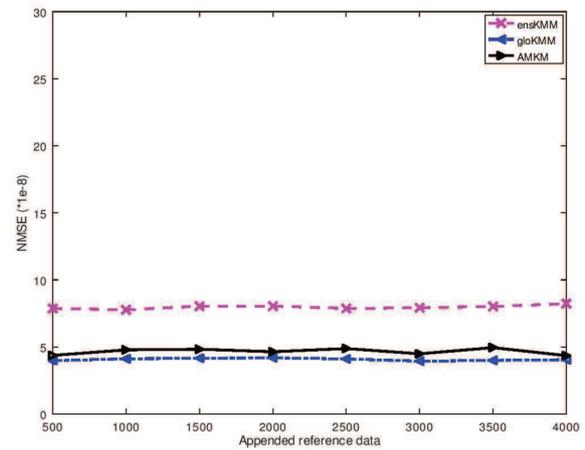}	}
    \caption{The experimental results of scalable learning on Forest and Letter data sets: (a) Forest and (b) Letter.}
\end{figure*}
\begin{table}[h]
	\centering
	\caption{The obtained average NMSE ($ \times 10^{-5}$ on Monks, Ionosphere, and Climate data sets. $ \times 10^{-7}$ on Forest, Letter and CIFAR data sets) from AMKM method with different quantities of randomly selected instances $ n $.}
	\begin{tabular}{|l||c|c|c|c|c|}
		\hline
		\multicolumn{2}{|c|} {Selected instances $ n $}   & 50 & 100 & 150 & 200	\\
		\hline\hline
		\multirow{3}{*}{Data sets}  
		& Monks & 1.059 & 1.121 & 1.254 & 1.204		\\
		& Ionosphere & 0.706 & 0.749 & 0.734 & 0.709 		\\
		& Climate & 1.538 & 1.418 & 1.702 & 1.463 		\\
		\hline\hline
		\multicolumn{2}{|c|} {Selected instances $ n $}   & 100 & 200 & 300 & 400	\\
		\hline\hline
		\multirow{3}{*}{Data sets}  
		& Forest & 1.804 & 2.006 & 2.124 & 2.278		\\
		& Letter & 0.502 & 0.509 & 0.535 & 0.528 		\\
		& CIFAR & 7.312 & 6.679 & 6.457 & 7.117 		\\
		\hline
	\end{tabular}
\end{table}
\begin{table}[h]
	\centering
	\caption{The average cost times (milliseconds) of AMKM with different quantities of randomly selected instances $ n $.}
	\begin{tabular}{|l||c|c|c|c|c|}
		\hline
		\multicolumn{2}{|c|} {Selected instances $ n $}   & 50 & 100 & 150 & 200	\\
		\hline\hline
		\multirow{3}{*}{Data sets}  
		& Monks & 63.031 & 65.618 & 71.402 & 74.994		\\
		& Ionosphere & 41.284 & 43.677 & 45.672 & 49.268 		\\
		& Climate & 54.647 & 58.836 & 62.427 & 65.619 		\\
		\hline\hline
		\multicolumn{2}{|c|} {Selected instances $ n $}   & 100 & 200 & 300 & 400	\\
		\hline\hline
		\multirow{3}{*}{Data sets}  
		& Forest & 68.741 & 121.4 & 199.192 & 264.816		\\
		& Letter & 77.311 & 117.211 & 191.407 & 258.434 		\\
		& CIFAR & 163.881 & 215.543 & 291.339 & 364.549 		\\
		\hline
	\end{tabular}
\end{table}
To exploit the influence of selected instances on AMKM, several experiments are performed to evaluate performance with different sizes of randomly selected instances. Among Monks, Ionosphere, and Climate data sets, 70 instances are randomly selected to be matching data and another 500, 250, 400 instances are respectively selected to be the reference data. For each experiment, the randomly selected instances of AMKM are set to be in range from 50 to 200, while top 100 important instances are used for final matching. Analogously, 500 instances are selected from Forest, Letter, and CIFAR data sets to be matching data, while 4,000 instances are selected to be reference data. During the experiments, the randomly selected instances of AMKM are set to be in range from 100 to 400, and the top 100 instances are adopted. The average experimental results of \emph{five} random repetitions are given in Tab. II. Obviously, the matching performance is almostly unaffected by changing sizes of selected random instances during AMKM approaches. In addition, the average time complexities of AMKM on each data set are shown in Tab. III. In terms of the experimental results, the time complexities of AMKM augment as the randomly selected instances are increased. In other words, the quantity of selected instances can be chosen as a regularly small size for efficiency, while matching performance is preserved.

Furthermore, another experiment tests the influence of most important instances for matching, and the same setting of matching and reference data are employed. Meanwhile, fixed 50 instances are randomly selected, and different quantities of important instances are selected for matching of gloKMM and AMKM. The obtained results are given in Tab. IV. With few instances, the results of gloKMM and AMKM are quite stable with different quantities of important reference data. On the contrary, the performance of gloKMM and AMKM get better as quantities of selected important instances are increased, which gives a positive influence on final results.
\begin{table}[h]
	\centering
	\caption{The obtained average NMSE ($ \times 10^{-5}$ on Monks, Ionosphere, and Climate data sets. $ \times 10^{-7}$ on Forest, Letter and CIFAR data sets) from AMKM method with different quantities of selected top important instances $ n_s $.}
	\begin{tabular}{|c|c|c|c|c|c|c|c|}
		\hline
		\multicolumn{2}{|c|} {Top instances $ n_s $}   & 50 & 100 & 150 & 200	\\
		\hline\hline
		\multirow{2}{*}{Monks} & gloKMM & 0.992 & 1.005 & 1.03 & 1.018		\\
		& AMKM & 1.249 & 1.209 & 1.056 & 1.076		\\
		\hline
		\multirow{2}{*}{Ionosphere} & gloKMM & 1.034 & 1.052 & 1.03 & 1.025 		\\
		& AMKM & 0.839 & 0.757 & 0.128 & 0.108		\\
		\hline
		 \multirow{2}{*}{Climate} & gloKMM & 1 & 1.051 & 1.026 & 1.001 		\\  
		& AMKM & 1.531 & 1.475 & 1.468 & 1.453		\\
		\hline\hline
		\multicolumn{2}{|c|} {Top instances $ n_s $}   & 100 & 200 & 300 & 400	\\
		\hline\hline
		\multirow{2}{*}{Forest} & gloKMM & 1.901 & 1.739 & 1.699 & 1.667		\\
		& AMKM & 1.782 & 1.49 & 1.353 & 1.381		\\
		\hline
		\multirow{2}{*}{Letter} & gloKMM & 0.412 & 0.394 & 0.393 & 0.385		\\
		& AMKM & 0.469 & 0.413 & 0.43 & 0.419		\\
		\hline
		\multirow{2}{*}{CIFAR} & gloKMM & 12.56 & 9.645 & 6.715 & 6.388		\\
		& AMKM & 9.074 & 7.741 & 5.93 & 5.897		\\
		\hline
	\end{tabular}
\end{table}

\section{Conclusion}
With the advances of information technologies, knowledge discovery and management have become more and more important in many real-world applications and intelligent systems. 
In the literature, kernel mean matching (KMM) based density ratio has been quite popular for its broad application characteristics,
but quite few available solutions to adaptive matching. In this work, a novel KMM method is proposed to adaptive learning of KMM.
More specifically, the importance estimation of instances are to be predicted with the density possibilities of instances and important samples are adopted for further steps of data handling by design.
As a consequence, calculation efficiency can be obtained with selective reference instances, and importance estimation of whole data can be avoided, which may suffer from the complexity problem of large data. 
Furthermore, scalable matching of kernel means can be conducted in the proposed method with selectively adaptive mechanism. Experimental results on a variety of data sets demonstrate that, the proposed method is able to obtain ideal KMM performance while promising efficiency can be achieved.

\section*{Acknowledgment}
The authors would like to thank 
anonymous referees for their 
constructive suggestions, and 
University of Toronto for providing CIFAR-100 data set publicly.




\bibliographystyle{IEEEtran}
\bibliography{./amkm.bib}
%

%
%

\end{document}